%% file: main.tex
\documentclass{article}

\PassOptionsToPackage{numbers, compress}{natbib}

\usepackage[final]{neurips_2024}

\usepackage[utf8]{inputenc} 
\usepackage[T1]{fontenc}    
\usepackage{hyperref}       
\usepackage{url}            
\usepackage{booktabs}       
\usepackage{amsfonts}       
\usepackage{nicefrac}       
\usepackage{microtype}      
\usepackage{xcolor}         

\usepackage{subcaption}
\usepackage{lineno}
\usepackage{tikz}
\usepackage{mathtools}
\usepackage{multirow}
\usepackage{helvet}
\usepackage{courier}
\usepackage[ruled]{algorithm2e}
\usepackage{amsmath,amssymb,mathrsfs,bm}
\usepackage{wrapfig}
\usepackage{pifont}
\usepackage{pgfplots}
\usepackage{caption}
\usepackage{paralist}
\usepackage{CJKutf8}
\usepackage{colortbl}
\usepackage{arydshln}
\usepackage{color}
\usepackage{CJKutf8}
\usepackage{enumitem}
\usepackage{makecell}
\usepackage{tabularx}
\usepackage{siunitx}

\newcommand{\bluehighlight}[1]{{\bf\color{blue}#1}}

\newcommand{\paratitle}[1]{\vspace{1.2ex}\noindent \textbf{#1}}

\title{Synergistic Dual Spatial-aware Generation\\ of Image-to-Text and Text-to-Image}

\author{%
  Yu Zhao$^1$, \, Hao Fei$^2$\thanks{Corresponding Author: Hao Fei}, \, Xiangtai Li$^{3}$, \, Libo Qin$^4$, \, Jiayi Ji$^2$, \\ \textbf{Hongyuan Zhu$^5$,\, Meishan Zhang$^6$, \, Min Zhang$^6$, \, Jianguo Wei$^1$}\\
  $^1$ Tianjin University\, $^2$ National University of Singapore\, $^3$ Nanyang Technological University\\ [2pt] $^4$ Central South University\,
   $^5$ I$^2$R \& CFAR,\,  A*STAR\thanks{Institute for Infocomm Research (I$^2$R) \& Centre for Frontier AI Research (CFAR), A*STAR} \quad  $^6$ Harbin Institute of Technology (Shenzhen)\\
  \texttt{zhaoyucs@tju.edu.cn, haofei37@nus.edu.sg}
}

\begin{document}

\maketitle

\begin{abstract}
In the visual spatial understanding (VSU) area, spatial image-to-text (SI2T) and spatial text-to-image (ST2I) are two fundamental tasks that appear in dual form.
Existing methods for standalone SI2T or ST2I perform imperfectly in spatial understanding, due to the difficulty of 3D-wise spatial feature modeling.
In this work, we consider modeling the SI2T and ST2I together under a dual learning framework.
During the dual framework, we then propose to represent the 3D spatial scene features with a novel 3D scene graph (3DSG) representation that can be shared and beneficial to both tasks.
Further, inspired by the intuition that the easier 3D$\to$image and 3D$\to$text processes also exist symmetrically in the ST2I and SI2T, respectively, we propose the Spatial Dual Discrete Diffusion (SD$^3$) framework, which utilizes the intermediate features of the 3D$\to$X processes to guide the hard X$\to$3D processes, such that the overall ST2I and SI2T will benefit each other.
On the visual spatial understanding dataset VSD, our system outperforms the mainstream T2I and I2T methods significantly.
Further in-depth analysis reveals how our dual learning strategy advances.
\end{abstract}

\section{Introduction}

Within the research topic of Visual Spatial Understanding (VSU) \citep{DBLP:conf/clef/KordjamshidiRMP17, DBLP:conf/emnlp/ZhaoWLSZ022, DBLP:journals/tcsv/SunZRWQL24, DBLP:conf/iccv/YangRD19,DBLP:journals/corr/abs-2401-12168}, Spatial Image-to-Text (SI2T) \citep{DBLP:conf/emnlp/ZhaoWLSZ022,DBLP:conf/acl/Zhao00WZZC23} and Spatial Text-to-Image (ST2I) \citep{qu2023layoutllm} are two representative task forms across vision and language.
SI2T aims to understand the spatial relationships of objects in the given image, while ST2I focuses on synthesizing a spatial-faithful image based on the input text prompts.
Existing efforts mostly formalize SI2T and ST2I tasks as the normal I2T and T2I problems, applying general-purpose I2T and T2I models for task solutions \citep{DBLP:conf/emnlp/ZhaoWLSZ022,DBLP:conf/acl/Zhao00WZZC23,qu2023layoutllm,ji2021improving,ji2022knowing}.
Technically, researchers widely employ the vision-language generative architecture \citep{DBLP:conf/icml/WangYMLBLMZZY22,DBLP:conf/icml/ChoLTB21,wu2023cross2stra} for I2T tasks, i.e., generally, there is a visual encoder and a text decoder.
For the T2I task, recent diffusion-based methods have shown the extraordinary capability of image generation and achieved state-of-the-art (SoTA) performance on a mount of benchmarks \citep{DBLP:conf/icml/Bar-TalYLD23, DBLP:conf/nips/DhariwalN21, DBLP:conf/cvpr/GuCBWZCYG22, DBLP:conf/iclr/SongME21, DBLP:conf/cvpr/RombachBLEO22,wu2024imagine,fei2024dysen,daneshfar2024image}.

Unfortunately, these strong-performing I2T and T2I methods, without the deliberate spatial semantics modeling in VSU tasks, largely fall short in precisely extracting spatial features from visual or textual inputs.
For instance, in the SI2T process, the spatial relationships are often incorrectly recognized due to layout overlap and perspective illusion \cite{DBLP:conf/acl/Zhao00WZZC23}.
This is due to the inherent characteristics of 2D images, which, lacking 3D feature modeling, inevitably fail to understand spatial relations.
On the other hand, in ST2I, synthetic images frequently fail to match strictly the spatial constraints specified in the prompts, such as the position, pose, and perspective \cite{qu2023layoutllm}.
This is because language's abstract nature allows only for a general description of the content rather than detailed depictions of spatial scenes. 
Moreover, unlike general image generation, ST2I should place greater emphasis on 3D spatial modeling, while the input sequential language tokens intrinsically do not portray the kind of specific spatial scene.

\begin{wrapfigure}[26]{r}{7cm}
\vspace{-3mm}
\centering
\begin{minipage}[t]{0.98\linewidth}
        \centering
        \includegraphics[width=\linewidth]{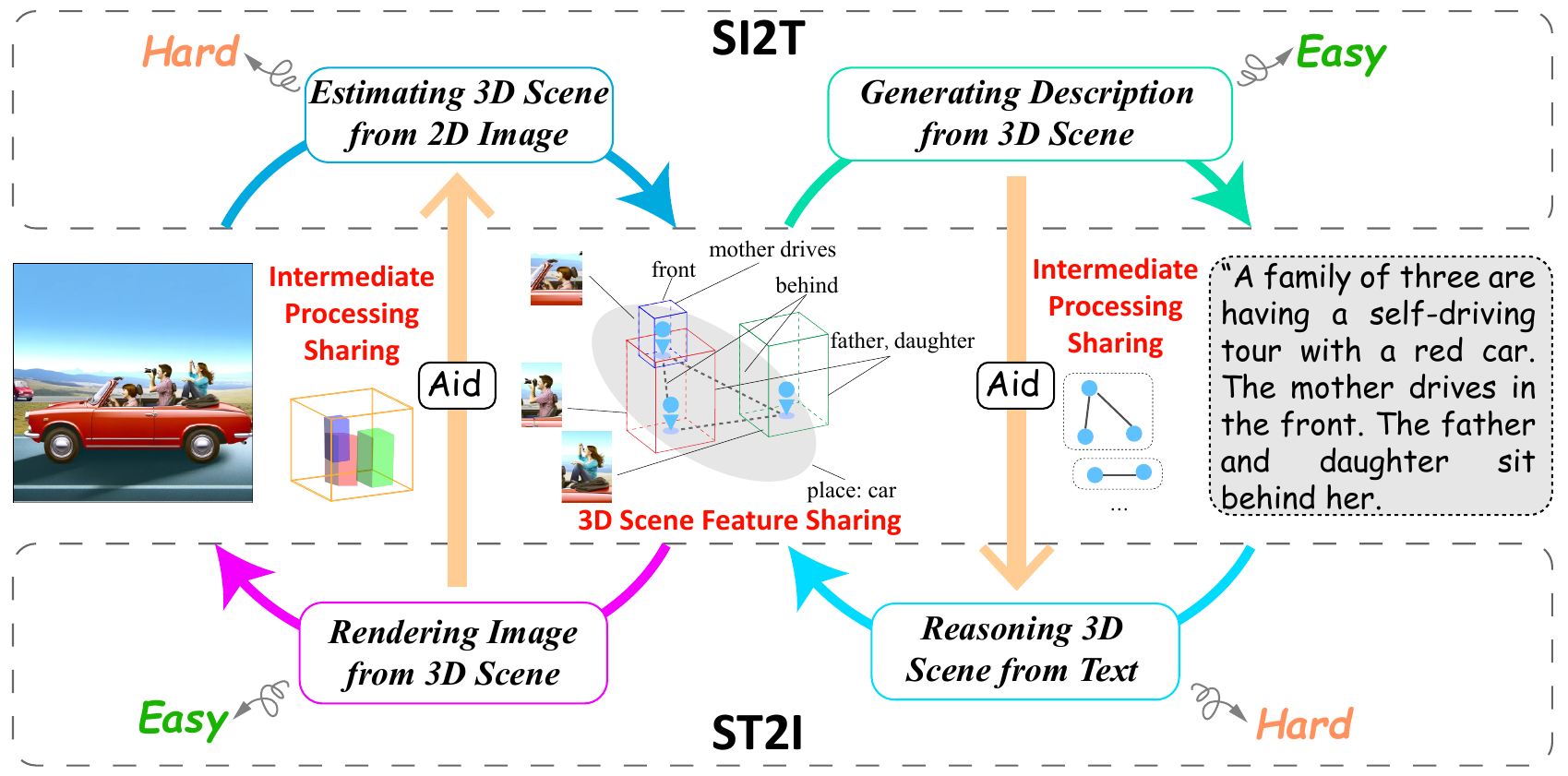}
        \caption*{\small (a) Complementary dual tasks of ST2I and SI2T.}
    \end{minipage}
    \begin{minipage}[t]{0.98\linewidth}
        \centering
        \includegraphics[width=\linewidth]{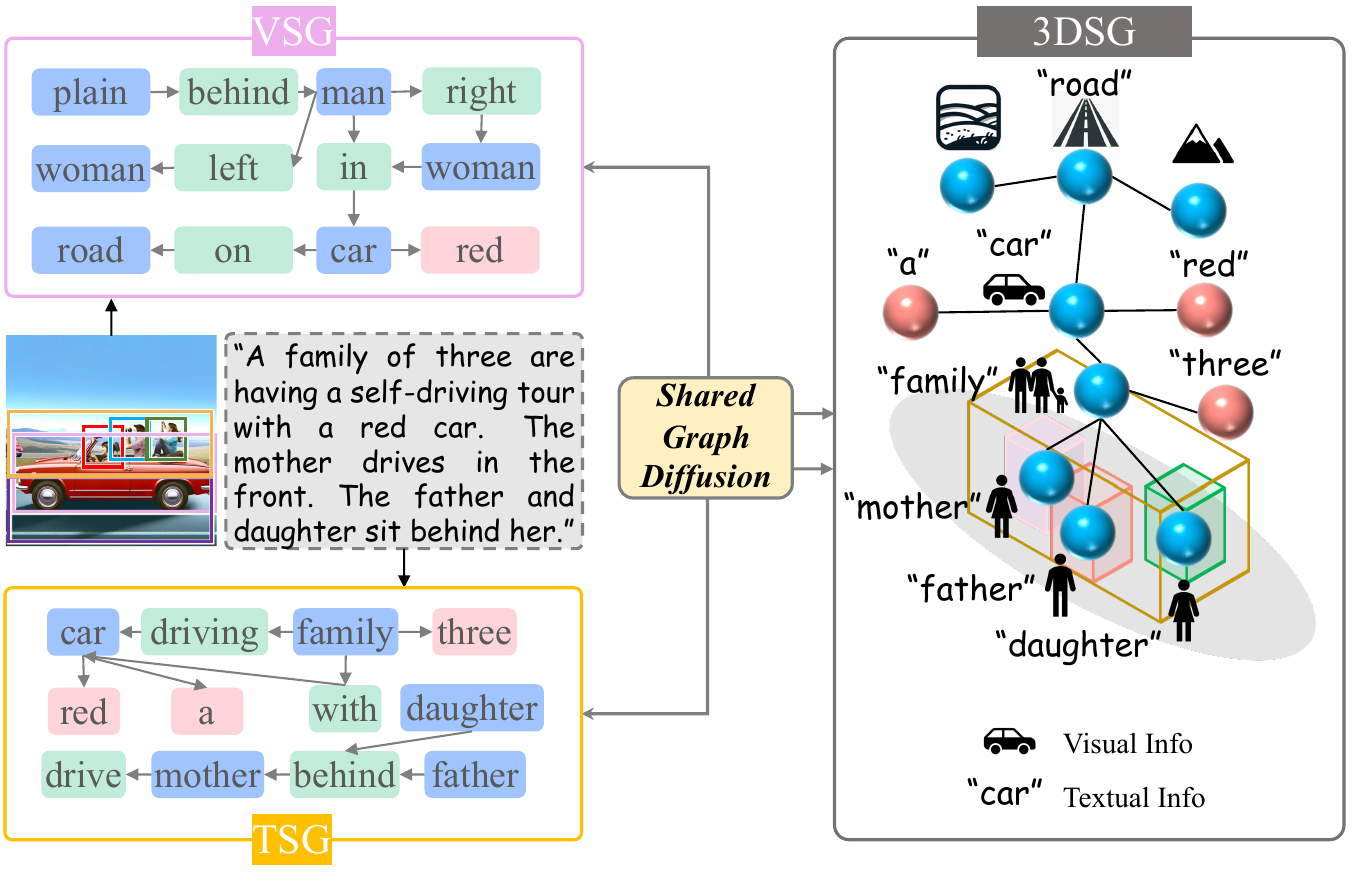}
        \caption*{\small (b) Diffusion Based 3DSG Generation.}
    \end{minipage}
    \vspace{-2mm}
\caption{Demonstration of SI2T and ST2I tasks.
}
\label{fig:intro}
\end{wrapfigure}
Let us revisit the human process in solving SI2T and ST2I tasks.
We typically process the input image or text within our minds by constructing a 3D spatial scene, which serves as the center for further spatial textual descriptions or image generation.
Specifically for SI2T, we intuitively project a 2D image into a reasonable 3D scene based on common sense before describing that scene in words. 
In contrast, for ST2I tasks, we start by imaginatively converting the text of user instruction into a 3D conceptual scene which is then rendered into a 2D image.
Interestingly, we can actually find that SI2T and ST2I are dual processes, with each task's input and output being the reverse of the other, as illustrated in Figure \ref{fig:intro}(a). 
More importantly, there are two important observations in these dual tasks.
\bluehighlight{First, [\emph{intermediate processing sharing}]}, they can complement and benefit each other. 
For SI2T, the `Image$\to$3D' reasoning process is challenging in acquiring necessary 3D features, whereas the descriptive `3D$\to$Text' process is relatively easier. 
Conversely, for ST2I, the `Text$\to$3D' process requires complex reasoning of the 3D scene feature, while rendering `3D$\to$Image' is much more straightforward.
Ideally, if complementing the information during each learning process, i.e., {letting the easy part aid the hard part}, it should enhance the performance of both tasks.
\bluehighlight{Second, [\emph{3D scene feature sharing}]}, both dual tasks urgently require modeling of the respective 3D features, where such stereospecific insights in 3D perspective can be essentially shared and also complementary between each other.
Intuitively, vision offers very concrete clues describing spatial scenes, e.g., objects and attributes, while the 3D features derived from texts are more likely to define the semantics-oriented relations or constraints, which are often abstract and rough.

Inspired by such intuition, in this paper, we introduce a novel synergistic dual framework for SI2T and ST2I.
To start with, we propose a spatial-aware 3D scene graph (namely \textbf{3DSG}), where the spatial objects, their relations and the layouts within the 3D scene are formulated with a semantically structured representation.
The {3DSG} representation effectively depicts the stereospecific attributes of all objects, and meanwhile models the spatial relations between them, from which both the SI2T and ST2I processes can be beneficial.
Technically, {3DSG} is obtained from a shared {graph diffusion} model for both SI2T and ST2I processes, as shown in Figure \ref{fig:intro}(b).
Trained with our `2DSG-3DSG' pair data, the graph diffusion is learned to propagate and evolve the initial 2D visual SG (parsed from input image at SI2T side) or the textual SG (parsed from input text at ST2I side) into the final {3DSG}, i.e., by adding all necessary and reasonable spatial details.

With \emph{3DSG}, we next implement the whole dual framework of SI2T and ST2I generation.
Instead of directly employing the SoTA generative I2T models or diffusion-based T2I methods, we consider a solution fully based on discrete diffusions \cite{DBLP:conf/nips/AustinJHTB21}, due to several key rationales.
Primarily, for VSU, the most crucial spatial information that determines a scene consists of objects and their relationships, which presents the characteristic of discretization and combination in the spatial layout, while other background and spatial unrelated information would be noisy.
Thus, the discrete representation is more appropriate to model pure spatial semantics in our scenario.
Besides, in our scenario both the textual token and SG representations possess discrete characteristics \cite{DBLP:journals/corr/abs-2403-16883}, which can be perfectly modeled by discrete diffusion.
Moreover, the discrete diffusion works on the limited index space \cite{DBLP:conf/cvpr/GuCBWZCYG22, DBLP:journals/corr/abs-2403-16883,DBLP:conf/cvpr/HuWCYS22, DBLP:journals/corr/abs-2211-11694}, which is much more computationally efficient, especially for visual synthesis tasks.
As illustrated in Figure \ref{fig:model}, our \textbf{S}patial \textbf{D}ual \textbf{D}iscrete \textbf{D}iffusion (dubbed as \textbf{SD}$\bm{^3}$) system consists of three components: 1) \emph{3DSG} generation, 2) SI2T generation and 3) ST2I generation.
During dual learning, both SI2T and ST2I first induce the \emph{3DSG} representations from the input image or text respectively via one shared graph diffusion, where the modality-variant features are simultaneously preserved in \emph{3DSG} for unbiased and holistic modeling of 3D spatial scene.
Then, the image synthesis (for ST2I) and text generation (SI2T) are carried out via two separate discrete diffusion models, during which the \emph{3DSG} feature is integrated for better generation.
At the meanwhile, the intermediate features of the `3D$\to$X' (X means text or image) diffusion steps are also passed to the counterpart hard `X$\to$3D' processes for further facilitation.

We conduct experiments on the VSD dataset \cite{DBLP:conf/emnlp/ZhaoWLSZ022}, which is a VSU benchmark with paired images and texts of spatial descriptions that allow for both SI2T and ST2I.
Extensive results demonstrate that our proposed system significantly outperforms all baselines on both ST2I and SI2T, including the vision-language model based and diffusion-based methods.
Further analysis reveals that the dual framework helps align the asymmetric spatial semantics across image and text modalities.
To our knowledge, this is the first attempt to resolve the SI2T and ST2I tasks with a novel dual perspective, and further successfully investigate synergistic learning in between by sufficiently modeling the spatial 3D scene representations.

\vspace{-3mm}
\section{Related Work}

\vspace{-3mm}
\paratitle{Visual Spatial Understanding.}
VSU is an important topic within the research of multimodal learning \cite{ma2022xclip,fei2023scene,fei2024video,fei2024enhancing}.
VSU aims to extract spatial information from a given scene,
developed within the forms of reasoning \cite{liu2023visual}, relation extraction \cite{DBLP:conf/mm/000223}, role labeling \cite{DBLP:conf/clef/KordjamshidiRMP17}, question answering \cite{DBLP:journals/corr/abs-2209-10326,DBLP:journals/corr/abs-2308-09778,DBLP:conf/naacl/MirzaeeFNK21}, image-to-text generation \cite{DBLP:conf/emnlp/ZhaoWLSZ022, DBLP:conf/acl/Zhao00WZZC23}, image synthesis \cite{qu2023layoutllm}, 3D reconstruction \cite{DBLP:journals/tcsv/SunZRWQL24,DBLP:conf/mmasia/ShaZPKY23}, etc.
With the VSU capability, many downstream applications can achieve,
such as robotics \citep{DBLP:journals/jair/FrancisKLLNO22,DBLP:journals/jair/MogadalaKK21}, navigation \citep{DBLP:conf/emnlp/HongOWG20,DBLP:journals/tie/WangQHYYG23,DBLP:journals/anor/BretasMJCSC23}, and language grounding \citep{DBLP:conf/acl/0032YF022}.
Among various VSU tasks, the SI2T and ST2I generation attract significant attention due to their fundamental positions in vision-language cross-modal tasks.
Current efforts mostly study ST2I or SI2T separately.
For the text-to-image generation, the diffusion models have emerged as the SoTA approaches \cite{DBLP:conf/aaai/FanC00YW23}.
For image-to-text generation, i.e., image captioning, it has been a long-standing task and has achieved great progress.
Recent advances can be largely attributed to vision-language pre-training (VLP) \cite{DBLP:conf/icml/WangYMLBLMZZY22, DBLP:conf/icml/ChoLTB21}.
Besides conventional VLP methods, \cite{DBLP:journals/corr/abs-2211-11694} are the first to use diffusion models for image captioning (and more specifically, for text generation).
In this work, we consider the two tasks together under a dual learning framework, via which we aim to achieve mutual benefits of the spatial feature modeling from each other.

\vspace{-2mm}
\paratitle{Discrete Diffusion Models.}
The diffusion model is first proposed by \cite{DBLP:conf/nips/HoJA20}, and has achieved impressive performance for text-to-image generation \cite{DBLP:conf/cvpr/RombachBLEO22,DBLP:conf/nips/DhariwalN21,DBLP:conf/iclr/SongME21,DBLP:conf/nips/SahariaCSLWDGLA22,wu2024imagine}.
Original diffusion models are parameterized Markov chains trained to translate simple distributions to more sophisticated target distributions in a finite set of steps on the continuous data or its latent representations.
In this work we adopt the SoTA diffusion-based model as our T2I backbone.
Recently, diffusion models on discrete space are introduced to markedly reduce the computing cost \cite{DBLP:conf/cvpr/GuCBWZCYG22,DBLP:conf/nips/AustinJHTB21,DBLP:journals/corr/abs-2102-05379,DBLP:conf/cvpr/HuWCYS22}.
Discrete diffusion methods are also used in text generation \cite{DBLP:journals/corr/abs-2102-05379} and structure generation \cite{DBLP:journals/corr/abs-2403-16883} due to its natural adaptability for the data with discrete nature.
This work follows the line and takes the SoTA discrete diffusion model as the backbone for image, text and scene graph generation.

\vspace{-2mm}
\paratitle{Scene Graph Representations.}
This work is also closely related to scene graph (SG)-based representation learning. 
Scene graphs have advanced in depicting the intrinsic semantic structures of scenes in images or texts \cite{DBLP:journals/ijcv/KrishnaZGJHKCKL17, Wang2018}. 
In SGs, key object and attribute nodes are connected via pairwise relations to describe semantic contexts, which have been shown to be useful as auxiliary features that carry rich contextual and semantic information for a wide range of downstream applications, such as image retrieval \cite{Johnson2015}, image generation \cite{Johnson2018,wu2024imagine}, translation \cite{fei2023scene}, image captioning \cite{Yang2019,wu2023cross2stra} and video modeling \cite{zhao2023constructing,fei2024dysen}. 
However, research on utilizing 3D scene graph representations to enhance various downstream tasks can be currently very scarce. 
In this paper, we incorporate both visual and language scene graphs within a 3D scope to enhance cross-modal alignment learning for better spatial semantics understanding.

\vspace{-2mm}
\paratitle{Dual Learning.} 
The dual learning method is proposed to enhance the coupled tasks that have the same exact input and output but in reverse \cite{DBLP:conf/icml/XiaQCBYL17,DBLP:conf/acl/YeLW19}.
With dual bidirectional learning, the model could capture potential mutual information from the primary and dual tasks, improving their performance.
Recently, the idea of dual learning has been applied to various tasks, such as intertranslation \cite{DBLP:conf/icml/XiaQCBYL17}, speech recognition with text-to-speech \cite{DBLP:conf/asru/TjandraS017}, question answering with question generation \cite{DBLP:journals/corr/TangDQZ17, DBLP:journals/tkde/SunTDQLYZLYFQL20}, and image classification with image generation \cite{DBLP:conf/icml/0001WRZ22}.
The key point of dual learning is to model the paired and complementary features between the coupled tasks, and then the learning process can reinforce them mutually.
In this work, we first connect the dual SI2T and ST2I tasks with their shared 3D spatial feature and use dual learning to enhance 3D feature construction.
To our knowledge, we are the first to achieve synergy between two spatial-aware cross-modal dual generations.

\vspace{-2mm}
\section{Preliminaries}

\vspace{-2mm}
\subsection{Task Formulation}
\vspace{-2mm}
We process the dual visual spatial understanding tasks, ST2I and SI2T.
Given a textual prompt $Y$, ST2I aims to generate an image $\hat{I}$ that semantically matches the spatial constraints with $Y$.
Its dual task is SI2T, which is also known as the visual spatial description (VSD), aiming to generate a piece of textual description $\hat{Y}$ based on an input image $I$.
In this paper, we process the two tasks parallelly.

\vspace{-2mm}
\subsection{Discrete Diffusion}
Diffusion models \cite{DBLP:conf/nips/HoJA20} are generative models characterized by a forward and reverse Markov process. 
In the forward process, the given data $\bm{x}_0$ with distribution $q(\bm{x}_0)$ is corrupted into a Gaussian distribution variable $\bm{x}_T$ in $T$ steps, formulated as $q(\bm{x}_{1:T}|\bm{x}_0)=\prod^{T}_{t=1}q(\bm{x}_t|\bm{x}_{t-1})$.
In the reverse process, the model learns to recover the original data $\bm{x}_0$ from $\bm{x}_T$, denoted as $p_{\theta}(\bm{x}_{0:T})=p(\bm{x}_T)\prod^{T}_{t=1}p_{\theta}(\bm{x}_{t-1}|\bm{x}_t)$.
In order to optimize the generative model $p_{\theta}(\bm{x}_0)$ to fit the data distribution $q(\bm{x}_0)$, one typically optimizes a variational upper bound on the negative log-likelihood:
\begin{equation}\label{eq:diff}\small
    \mathcal{L}_{vlb} = \mathbb{E}_{q(\bm{x}_0)}\Bigg[D_{KL}\left[q(\bm{x}_T|\bm{x}_0)||p(\bm{x}_T)\right]+\sum^T_{t=1}\mathbb{E}_{q(\bm{x}_t|\bm{x}_0)}\left[D_{KL}\left[q(\bm{x}_{t-1}|\bm{x}_t,\bm{x}_0)||p_{\theta}(\bm{x}_{t-1}|\bm{x}_t)\right]\right]\Bigg].
\end{equation}

Vanilla diffusion models are defined on continuous space.
Recently, the discrete diffusion model has been introduced where a transition probability matrix $Q_t$ is defined to indicate how $\bm{x}_0$ transits to $\bm{x}_t$ for each step of the forward process, where $\bm{x}_t\in \mathbb{Z}^{N}$ is defined in discrete space. 
The matrices $[Q_t]_{ij}=q(\bm{x}_t=i|\bm{x}_{t-1}=j)$ defines the probabilities that $\bm{x}_{t-1}$ transits to $\bm{x}_t$.
Then the forward and reverse process could be rewritten as $q(\bm{x}_t|\bm{x}_{t-1})=\bm{v}^{\top}(\bm{x}_t)Q_t\bm{v}(\bm{x}_{t-1})$ and $q(\bm{x}_{t-1}|\bm{x}_t,\bm{x}_0)=q(\bm{x}_t|\bm{x}_{t-1},\bm{x}_0)q(\bm{x}_{t-1}|\bm{x}_0)/q(\bm{x}_t|\bm{x}_0)$, where $\bm{v}(\bm{x})$ means the one-hot representation of $\bm{x}$.
Appendix \ref{app:ddiff} gives more technical details about the discrete diffusion models.

\input{tables/3DSG}
\vspace{-2mm}
\subsection{Scene Graph Representation}
\vspace{-2mm}
The scene graph presents a scene via the objects with their attributes and relationships in the form of a graph, which can be constructed from either text (TSG) \cite{DBLP:conf/acl-vl/SchusterKCFM15} or image (VSG) \cite{DBLP:conf/cvpr/ZellersYTC18,10447193,Li_Ji_Wu_Li_Qin_Wei_Zimmermann_2024}.
We represent a scene graph as $G=\{V,E\}$, where $V$ is the node set, and $E$ is the edge set.
There are three types of nodes in the scene graph: object nodes, attribute nodes, and relationship nodes.
Each type of node has ts own unique tag vocabulary.

For SI2T and ST2I, we focus on spatial information extraction, especially the object relationships in the 3D space.
However, due to the limited information presented by text or image, a 2D TSG or VSG hardly fully models these 3D spatial semantics.
Here, we introduce the spatial-aware 3D scene graph (3DSG) \cite{DBLP:journals/tcyb/KimPSK20}, which thoroughly depicts 3D scenes.
The 3DSG is formally equivalent to the 2D scene graph, i.e. the object, attribute, and relationship, but organized hierarchically, which contains more abundant elements such as nodes for high-level spatial concepts, texture attributes, and spatial relationships in the 3D perception.
Table \ref{tab:3DSG} presents the differences between 2DSG and our 3DSG.
Conventionally, the 3DSG should be constructed directly from a 3D scene, i.e., sensor data \cite{DBLP:journals/ijrr/RosinolVAHCSGC21}, point clouds \cite{DBLP:conf/cvpr/WaldDNT20}, 3D mesh \cite{DBLP:conf/iccv/ArmeniHZGMFS19}, and RGB-D sequences \cite{DBLP:conf/cvpr/WuWTNT21, DBLP:conf/nips/GothoskarCZGPGT21}.
In this work, we uniform the VSG/TSG/3DSG representations, modeling the object nodes $v_{obj}$, attribute nodes $\bm{v}_{attr}$, and relationship nodes $\bm{v}_{rel}$ by the embedding of their textual tags $\bm{e}_{obj},\bm{e}_{attr},\bm{e}_{rel} \in \mathbb{R}^{d}$, where $d$ is the dimension of the tag embedding.

\definecolor{mygreen}{RGB}{137,252,178}
\definecolor{myred}{RGB}{242,186,161}

\begin{figure*}[t]
\centering
\includegraphics[width=\linewidth]{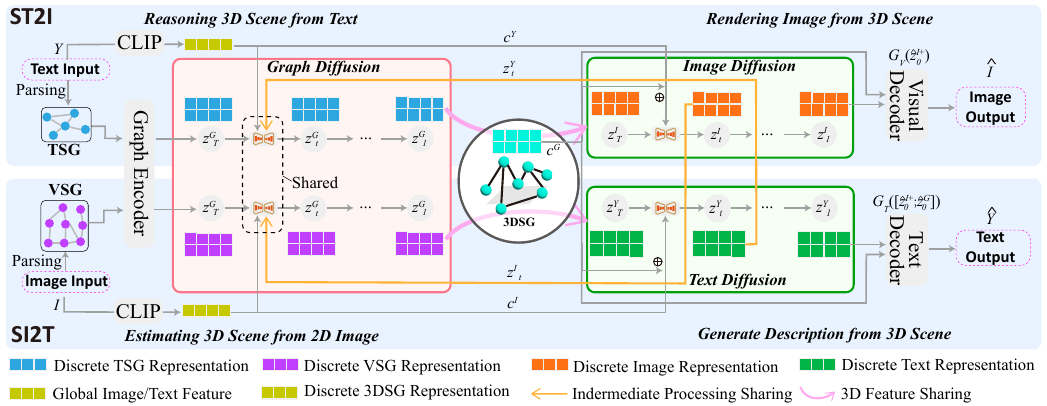}
\caption{Overall Framework of the S$^3$D. The figure presents the dual processes of ST2I and SI2T. The \textcolor{myred}{RED} block represents the hard X$\to$3D processes, and the \textcolor{mygreen}{GREEN} block represents the 3D$\to$X processes.
There are three diffusion processes in total, i.e., a shared graph diffusion model for VSG/TSG$\to$3DSG generation, and the image diffusion model and text diffusion model.
}
\vspace{-3mm}
\label{fig:model}
\end{figure*}

\vspace{-1mm}
\section{Methodology}

\vspace{-1mm}

\subsection{Spatial Dual Discrete Diffusion Framework}\label{sec:4.1}
In Figure \ref{fig:model}, we illustrate the overall framework of the proposed Spatial Dual Discrete Diffusion (SD$^3$), consisting of three separate discrete context diffusion models, i.e., the 3DSG diffusion model for the \textbf{X$\to$3D} process, the ST2I diffusion model for the \textbf{3D$\to$Image} process, and the SI2T diffusion model for the \textbf{3D$\to$Text} process.

\paratitle{X$\to$3D: 3DSG Diffusion Model.}
We set a graph diffusion model to convert the initial TSG (for ST2I) and VSG (for SI2T) to the 3DSG.
We first acquire the initial VSG and TSG from the input image and textual description through the off-the-shelf VSG \cite{DBLP:conf/acl-vl/SchusterKCFM15} and TSG \cite{DBLP:conf/cvpr/ZellersYTC18} parsers.
Then the initial VSG and TSG nodes (represented as $\{\bm{e}_{obj},\bm{e}_{attr}, \bm{e}_{rel}\}$) are encoded to the latent quantized representations by a discrete graph auto-encoder (DGAE) \cite{DBLP:journals/corr/abs-2403-16883}, on which the discrete diffusion process on discrete graph representations is conducted to generate the 3DSG.
Concretely, we denote the quantized gold 3DSG as $\bm{z}^G_0$.
Then, following conventional discrete diffusion, we can calculate $p^{Graph}_{\psi}(\bm{z}^G_{t-1}|\bm{z}^G_t)$.
Moreover, on our final dual training, the 3DSG diffusion model leverages the intermediate feature of the following \textbf{3D$\to$X} process to aid the  \textbf{TSG$\to$3DSG} and \textbf{VSG$\to$ 3DSG} generation process.
For \textbf{TSG$\to$3DSG}, 
the model takes global text features and the intermediate feature of the dual SI2T diffusion, i.e., $p^{Graph}_{\psi}(\bm{z}^G_{t-1}|\bm{z}^G_t, \bm{c}^Y)$, $p^{Graph}_{\psi}(\bm{z}^G_{t-1}|\bm{z}^G_t, \bm{z}^Y_t)$, and for \textbf{VSG$\to$3DSG} it will be $p^{Graph}_{\psi}(\bm{z}^G_{t-1}|\bm{z}^G_t, \bm{c}^I)$, $p^{Graph}_{\psi}(\bm{z}^G_{t-1}|\bm{z}^G_t, \bm{z}^I_t)$, where the $\bm{z}^Y_t$ and $\bm{z}^I_t$ are the $t$ step representations of SI2T and ST2I diffusion respectively.
Then we align them by:
\begin{align}\small
    &\mathcal{L}_{X-T23D} = D_{KL}(p^{Graph}_{\psi}(\bm{z}^G_{t-1}|\bm{z}^G_t, \bm{c}^Y)||p^{Graph}_{\psi}(\bm{z}^G_{t-1}|\bm{z}^G_t, \bm{z}^Y_t)),\\
    &\mathcal{L}_{X-I23D} = D_{KL}(p^{Graph}_{\psi}(\bm{z}^G_{t-1}|\bm{z}^G_t, \bm{c}^I)||p^{Graph}_{\psi}(\bm{z}^G_{t-1}|\bm{z}^G_t, \bm{z}^I_t)).
\end{align}

To enable the training for this 2DSG$\to$3DSG process, we should construct the paired `2DSG-3DSG' data.
We follow previous work \cite{DBLP:journals/corr/abs-2312-11713,DBLP:conf/cvpr/WaldDNT20,DBLP:journals/corr/abs-2309-06635} to construct a gold 3DSG dataset.
After that, we adopt a strong captioning model to generate text descriptions from view images. 
With all the inputs and outputs ready, we can train the graph diffusion model on aligned 3DSG-Image-Text data, after which the model will acquire the 3D estimation capability.

\paratitle{3D$\to$Image: ST2I Diffusion Model.}
The ST2I diffusion model is used to generate image $\hat{I}$ with the condition of input textual prompts $Y$.
In this task, we adopt a vector quantized variational autoencoder (VQ-VAE) \cite{DBLP:conf/cvpr/EsserRO21} as the quantized model to encode image data to embedding vectors.
We denote the quantized image as $\bm{z}^{I}_0$ and we could also calculate $q(\bm{z}^I_t|\bm{z}^I_{t-1})$ and $q(\bm{z}^I_{t-1}|\bm{z}^I_t, \bm{z}^I_0)$.
On the other hand, the ST2I diffusion model takes two conditions, i.e., the global text feature $\bm{c}^Y$ extracted via a CLIP model \cite{DBLP:conf/icml/RadfordKHRGASAM21}, and the shared 3DSG feature $\bm{c}^G$.
Then $\bm{c}^Y$ and $\bm{c}^G$ are fused and fed to the diffusion model to calculate $p^{ST2I}_{\theta}(\bm{z}^I_{t-1}|\bm{z}^I_t, \bm{c}^T\oplus\bm{c}^G)$.
At last, the decoder from VQ-VAE is used to generate images from latent codes.

\paratitle{3D$\to$Text: SI2T Diffusion Model.}
In the SI2T model, the inputs and outputs are reversed.
The diffusion is applied on the text representation $\bm{z}^Y_0$ with a similar process as ST2I.
The latent text representation is the word embedding, which is functionally equivalent to the visual codebook.
We use another denoising network $p^{SI2T}_{\phi}(\bm{z}^Y_{t-1}|\bm{z}^Y_t, \bm{c}^I\oplus\bm{c}^G)$, while the input condition becomes the visual tokens $\bm{c}^I$ and also the 3DSG feature $\bm{c}^G$.
Afterward, the language model decoder generates textual results based on the latent representations.

\vspace{-2mm}
\paratitle{Image and Text Decoding.}
Finally, through three diffusion models, we acquire the predicted latent representations of image, text, and 3DSG, denoted as $\bm{\hat{z}}^I_0$, $\bm{\hat{z}}^Y_0$, $\bm{\hat{z}}^G_0$.
For ST2I, we fuse the generated graph vectors $\bm{\hat{z}}^G_0$ to visual vectors $\bm{\hat{z}}^I_0$ by adopting an attention mechanism:
\begin{equation}
    Attn_{i,m} =\mathop{softmax}\limits_m(\bm{\hat{z}}^G_0[m] \cdot \bm{\hat{z}}^I_0[i]),\quad
    \bm{\hat{z}}^{I+}_0[i] = \bm{\hat{z}}^I_0[i]) \oplus \sum_{m}Attn_{i,m}\cdot \bm{\hat{z}}^G_0[m],
\end{equation}
where $\oplus$ means the concatenation operation.
Afterwards, the image decoder $G_V$ is used to reconstruct image from the scene graph enhanced codes $\bm{\hat{z}}^{I+}$:
\begin{equation}\label{eq:9}
    \hat{I} = G_V(\bm{\hat{z}}^{I+}).
\end{equation}
For SI2T, we append the flatten $\bm{\hat{z}}^G_0$ after generated textual codes $\bm{\hat{z}}^Y_0$ and use a language model decoder $G_T$ to generate the description:
\begin{align}\label{eq:10}
    \hat{Y} = G_T([\bm{\hat{z}}^Y_0;\bm{\hat{z}}^G_0]).
\end{align}

\vspace{-2mm}
\subsection{Mutual Spatial Synergistic Dual Learning}

\vspace{-2mm}
To exploit the complementarity of the dual processes, we elaborate a dual training strategy for our framework.
To conduct the strategy effectively, we introduce the essential training objectives.

\paratitle{Dual Learning Objective.}
Our dual learning framework contains two tasks, i.e. ST2I and SI2T, where their inputs and outputs are just reversed.
We denote ST2I as $f_{\theta}: i\to y, i\in I, y\in Y$ and SI2T as $f_{\phi}: y\to i , i\in I, y\in Y$.
Their learning objectives should be:
\begin{equation}    
    \mathcal{L}_{ST2I} = \mathbb{E}_{i,y}\log p_{\theta}(y|i), \quad
    \mathcal{L}_{ST2I} = \mathbb{E}_{i,y}\log p_{\phi}(y|i).
\end{equation}
Based on the dual supervising learning \cite{DBLP:conf/icml/XiaQCBYL17}, given the duality of the two tasks, if the learned ST2I and SI2T models are perfect, we should have the probabilistic duality:
\begin{equation}
    p(i)p_{\theta}(y|i)=p(y)p_{\phi}(i|y)=p(i,y), \forall i,y,
\end{equation}
where $p(i)$ and $p(y)$ are the marginal distributions.
Then we add this constraint to the dual learning target as an equivalent regularization term:
\begin{equation}\label{eq:dual}
    \mathcal{L}_{dual} = \mathcal{L}_{ST2I} + \mathcal{L}_{SI2T} + ||\log\hat{p}(i)+\log p_{\theta}(y|i)-\log\hat{p}(y)- \log p_{\phi}(i|y)||,
\end{equation}
where $\hat{p}(i)$ and $\log\hat{p}(y)$ are the estimated marginal distribution by a pre-trained vision-language model, which is illustrated detailly in Appendix \S\ref{app:dual}.

\paratitle{Loss for Mutual 3D Feature Learning.}
The 3DSG diffusion model is used to estimating 3D scene information from TSG or VSG, denoted as $f_{\psi}: i\to g$ and $f_{\psi}: y\to g$, where $i\in I, y\in Y, g\in G$.
Based \S\ref{sec:4.1} and Eq. \ref{eq:diff}, the graph diffusion loss can be acquired, denoted as $\mathcal{L}_{I23D}$ and $\mathcal{L}_{T23D}$
\begin{equation}    
    \mathcal{L}_{I23D} = \mathbb{E}_{i,y}\log p_{\theta}(g|i), \quad
    \mathcal{L}_{T23D} = \mathbb{E}_{i,y}\log p_{\phi}(g|y),
\end{equation}
where $z^I$ and $z^Y$ are the intermediate features of image diffusion and text diffusion.
Along with the X$\to$3D alignment loss $\mathcal{L}_{X-T23D}$ and $\mathcal{L}_{X-I23D}$, the final mutual 3D Feature learning target is:
\begin{equation}
    \mathcal{L}_{mutual} = \mathcal{L}_{I23D} + \mathcal{L}_{X-I23D} + \mathcal{L}_{T23D} + \mathcal{L}_{X-T23D}.
\end{equation}

\paratitle{Loss for Spatial Feature Alignment.}
The paired latent visual and textual representations are initialized in different feature spaces.
We thus conduct a spatial feature alignment to bridge them by the shared spatial representation of the 3DSG.
Concretely, we consider the image decoder, text decoder and graph encoder.
Given the paired image $I$, text $Y$ and the corresponding 3DSG $G_{host}$, we can easily get their quantified representations $z^I_0$, $z^Y_0$, $z^G_0$, and then the fused feature $z^{I+}_0$ and $[z^Y_0;z^G_0]$ based on the Eq. \ref{eq:9} and Eq. \ref{eq:10}.
After that, we adopt the reconstruct loss of VQ-VAE $\mathcal{L}_{v-dec}$,  and the next token prediction loss $\mathcal{L}_{t-dec}$ of text decoder for optimization:
\begin{equation}\label{eq:align}
    \mathcal{L}_{v-dec} = ||I - G_V(z^{I+}_0)||^2,\quad
    \mathcal{L}_{t-dec} = -\sum_{i=1}^{|y|}\log p(y_i|y_{<i},[z^Y_0;z^G_0]).
\end{equation}

\paratitle{Loss for 3DSG Reconstruction.}
To initialize the graph encoder and decoder, i.e., the DGAE, we follow \cite{boget2024discrete} to calculate the graph reconstruction loss:
\begin{equation}
    \mathcal{L}_{DGAE}=-\mathbb{E}_{\hat{\mathcal{Z}}^G} \ln (p(G_{host}|\hat{\mathcal{Z}}^G)),
\end{equation}
where $\hat{\mathcal{Z}}^G$ is the latent representations of the predicted 3DSG and the $G_{host}$ is the gold 3DSG.

\begin{wrapfigure}[22]{r}{7cm}
\centering
\includegraphics[width=\linewidth]{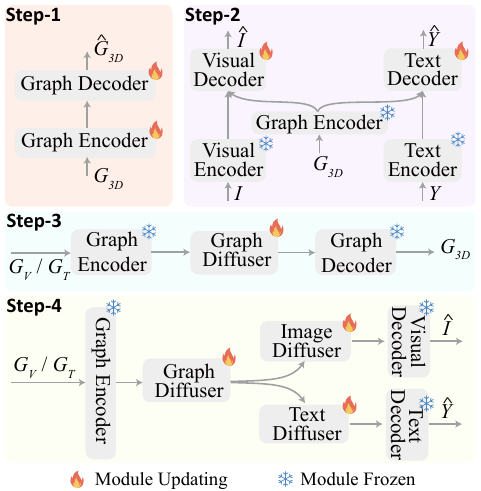}
\caption{Illustrations of the training steps of SD$^3$.}
\label{fig:training}
\end{wrapfigure}
\paratitle{Training Remarks.}
To achieve a perfect convergency, we use a four-step strategy for the overall training process, shown in Figure \ref{fig:training}.
\textbf{Step-1 DGAE pre-training.} First, we pre-train the DGAE model with the 3DSG reconstruction target $\mathcal{L}_{DGAE}$ to initialize the DGAE and the graph codebook, with the 
self-supervised training on gold 3DSG data.
For this step, we specially choose the 3DSG datasets \cite{DBLP:journals/corr/abs-2312-11713,DBLP:conf/cvpr/WaldDNT20,DBLP:journals/corr/abs-2309-06635} for training and the graph codebook will be well initialized.
\textbf{Step-2 Spatial Alignment.} Then we tune the visual and text decoder with the loss in Eq. \ref{eq:align}. During this step, only visual decoder and text decoder are updated with the loss $\mathcal{L}_{v-dec}+\mathcal{L}_{t-dec}$.
\textbf{Step-3 2DSG$\to$3DSG Diffusion Training.}
In this step, we use the constructed aligned 3DSG-Image-Text data (\S\ref{sec:4.1}) to train the 2DSG$\to$3DSG graph diffusion, thus the model will acquire the prior 3D estimation capability.
The optimization objectives here is $\mathcal{L}_{I23D}+\mathcal{L}_{T23D}$.
\textbf{Step-4 Overall Training.} Finally we tune the whole model with dual learning objectives, where we update the three diffusion models and freeze other modules.
The overall learning target is $\mathcal{L}_{dual} + \mathcal{L}_{mutual}$.

After above training steps, the SI2T or ST2I can be launched alone without the intermediate process sharing.
At this point, the critical graph diffusion module has learned sufficiently well from the previous training, so without the dual task aiding, the 3DSG generated during inference is also of high quality for the following 3D$\to$X generation.

\section{Experiment}

\vspace{-2mm}
\subsection{Experiment Settings}

\vspace{-2mm}
\paragraph{Dataset.}
To demonstrate the capability of our proposed method for both ST2I and SI2T generation, we conduct experiments on the VSD \cite{DBLP:conf/emnlp/ZhaoWLSZ022, DBLP:conf/acl/Zhao00WZZC23} dataset, which is constructed for visual spatial understanding.
The VSD dataset contains about 30K images from SpatialSense \cite{DBLP:conf/iccv/YangRD19} and Visual Genome \cite{DBLP:journals/ijcv/KrishnaZGJHKCKL17}, and each image in this dataset has aligned text for spatial scene description.
Now, VSD has two released versions, VSDv1 and VSDv2, where the spatial descriptions in VSDv2 are more meticulous.

\vspace{-2mm}
\paragraph{Evaluation metrics.}
For spatial text-to-image synthesis, following prior work, we adopt Fr\'{e}chet Inception Distance (FID) \cite{DBLP:conf/nips/SalimansGZCRCC16}, Inception score (IS) \cite{DBLP:conf/nips/HeuselRUNH17}, and CLIP score \cite{DBLP:conf/emnlp/HesselHFBC21} to evaluate authenticity of generated images.
For spatial image-to-text, we follow \cite{DBLP:conf/emnlp/ZhaoWLSZ022} and adopt BLEU4 \cite{DBLP:conf/acl/PapineniRWZ02} and SPICE \cite{DBLP:conf/eccv/AndersonFJG16} to measure text generation.
Furthermore, we perform human evaluation to compare the spatial understanding effectiveness.
Appendix \ref{app:ext_setting} details more experiment settings.

\vspace{-2mm}
\paragraph{Implementation Details.}
We use the pre-trained VQ-VAE of VQ-GAN \cite{DBLP:conf/cvpr/EsserRO21}, which leverages the GAN loss to get a more realistic image.
It converts 256×256 images into 32×32 tokens.
We follow \cite{DBLP:conf/cvpr/GuCBWZCYG22} and remove the useless codes and obtain a codebook with size K=2886.
For text encoder, we adopt the CLIP model and its tokenizer, which has 77 as the max token lengths.
We adopt the pre-trained GPT-2 \cite{radford2019language} as the text decoder, following its default settings.
We freeze the VQ-VAE encoder and CLIP while training.
For the user Unet to learn the denoising process for all the three diffusion models.
We follow the default settings of DGAE \cite{DBLP:journals/corr/abs-2403-16883} for the graph diffusion model.
We optimize the framework using AdamW \cite{DBLP:conf/iclr/LoshchilovH19} with $\beta_1$ = 0.9 and $\beta_2$ = 0.98. The learning rate is set to 5e-5 after 10,000 warmup iterations in the final dual tuning.

\input{tables/main_res}
\vspace{-1mm}
\subsection{Quantitative Results}
\vspace{-1mm}
\paragraph{Main Comparisons.}
The main results are shown in Table \ref{tab:result}
We first compare some strong baselines of I2T and T2I methods.
Overall, the ST2I results on VSDv2 are consistently better, and the SI2T results on VSDv1 are better.
This is because the VSDv2 contains more complicated and detailed textual descriptions, which is beneficial to ST2I but challenging for ST2I.
We see that the proposed SD$^3$ substantially outperforms the compared baselines on both VSDv1 and VSDv2.
For the ST2I task, we outperform Frido by 1.82\% FID, 3.28\% IS. and 3.66\% CLIP score on VSDv1 and 1.32\% FID, 3.74\% IS and 4.09\% CLIP score on VSDv2
For the SI2T task, we surpass 3DVSD by 1.38\% BLEU4 and 0.74\% SPICE on VSDv1 and 1.23\% BLEU4 and 1.06\% SPICE on VSDv2.
The results directly demonstrate the efficacy of our method.

We further show the key module ablation studies on the last four lines in Table \ref{tab:result}.
First, removing the 3DSG integration (``Vanilla Dual Learning''), i.e., train the SI2T and ST2I diffusion with loss of Eq. \ref{eq:dual}, the performance drops much on main metrics, indicating the critical influence of the 3DSG guidance.
Also, the drops of ``Singleton+3D'' reveal the effectiveness of our dual learning strategy.
When removing both 3DSG modeling and dual learning (``Singleton''), the model decay to the simple discrete diffusion model, which is just comparable to the VQ-Diffusion and DDCap model.

\input{tables/human}
\vspace{-2mm}
\paragraph{Spatial Evaluation.}
To compare the model's effectiveness, we conduct human evaluations for both ST2I and SI2T.
We follow \cite{DBLP:conf/acl/Zhao00WZZC23} to ask ten volunteers to answer the 5-point Likert scale on 100 samples.
For SI2T, we collect scores for \textit{Spatial Description Accuracy} and \textit{Spatial Description Diversity}.
For ST2I, we collect scores for \textit{Visual Spatial Accuracy}.
The results are shown in Table \ref{tab:human}.
Overall, our SD$^3$ method shows the best spatial understanding capability on SI2T and ST2I.
More details are put in Appendix \S\ref{app:ext_setting}.

\begin{figure}
    \centering
    \includegraphics[width=1\linewidth]{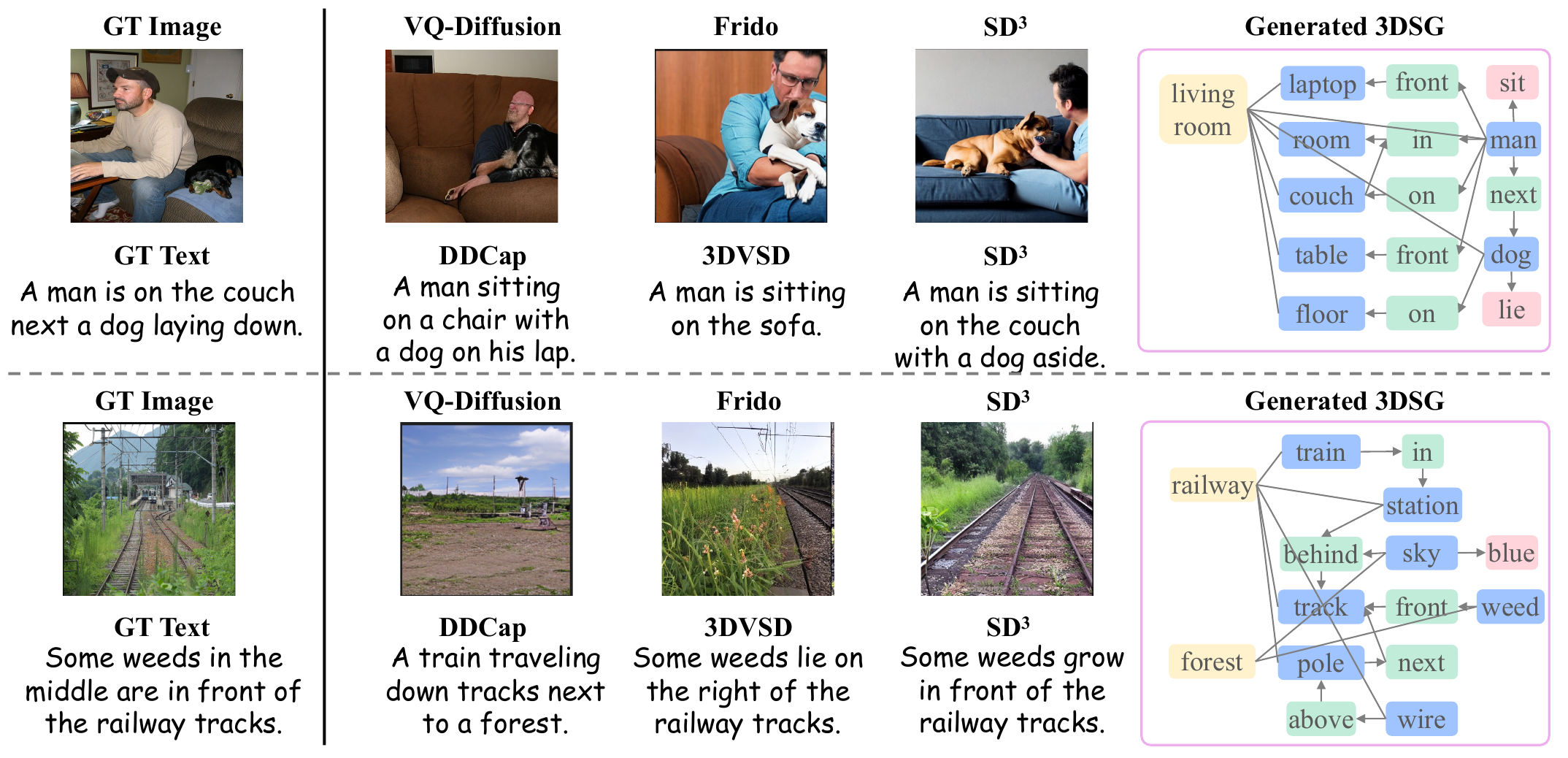}
    \vspace{-2mm}
    \caption{Qualitative results by different models, where the samples are selected from VSDv2.}
    \label{fig:case}
    \vspace{-5mm}
\end{figure}

\vspace{-2mm}
\subsection{Qualitative Results}
\vspace{-2mm}
To present the ST2I and SI2T performance of our model in a more intuitive way, we show some qualitative results in Figure \ref{fig:case}.
We visualize the generated images and texts using different methods.
For ST2I, compared with Frido and VQ-Diffusion, our SD$^3$ generates more spatial-faithful images.
For SI2T, compared with 3DVSD and DDCap, our method is able to extract and correctly describe the key spatial information.

\input{tables/ana3}

\vspace{-2mm}
\subsection{In-depth Analyses and Discussions}
\vspace{-2mm}
Previously, we have verified the effectiveness of our dual learning by thorough numerical evaluations.
Now, we explore how our methods advance via the following research questions.

\paratitle{How does 3DSG guidance aid the generation of spatial-faithful image and text?}
First, after removing the 3DSG integration, comparing the last two lines in Table \ref{tab:result}, we can see that the 3DSG feature contributes great influence.
Further, we follow GLIP \cite{DBLP:conf/cvpr/LiZZYLZWYZHCG22} to assess how well the objects-attribute matching between input image/text and generated ones, 
Also, we assess how well the subject-predicate-object triplets of the gold TSG and VSG could be retrieved in the ones from the generated image/text via the triplets recall (TriRec.) between scene graphs.
We compare the SoTA SI2T and ST2I baselines and the results are shown in Figure \ref{fig:q1}.
We observe that our SD$^3$ significantly outperforms the model without 3DSG guidance (``Vanilla Dual'') on the two metrics, which is just comparable to the baseline models.
This suggests that with the 3DSG guidance, the model could capture correct and sufficient spatial structures from inputs for the generation.

\paratitle{How does the dual intermediate sharing aid the graph diffusion model?}
On one hand, we report the performance of 3DSG generation on the gold 3DSG dataset \cite{DBLP:conf/cvpr/WaldDNT20} to evaluate how well the structures of generated 3DSG could be retrieved in the gold ones.
As shown in Table \ref{tab:q2}, we can see that compared with the initial VSG/TSG the generated 3DSG of our full model highly matches to the ground truth while without the dual intermediate sharing (``w/o Xfeat.''), i.e., remove the $\mathcal{L}_{X-I23D}$ and $\mathcal{L}_{X-T23D}$ in the \textbf{X$\to$3D} diffusion, the performance dramatically drops, showing the effectiveness of intermediate process sharing.
Besides, if the graph diffusion pre-training is removed (``w/o DiffPre.''), it will lose the ability to generate 3DSG (degrade to initial TSG/VSG).
On the other hand, we assess how much the graph diffusion model produce the new structures, compared with the initial TSG/VSG.
As shown in Figure \ref{fig:q2_diff}, after graph diffusion, the average numbers of the three types of elements are enriched substantially.

\input{figures/obj_ablation}

\paratitle{The influence of the training strategy.}
To show the effectiveness of our four-step training strategy, here we conduct an objective ablation study.
We investigate the model without DGAE pre-training in step 1, the model without spatial alignment in step 2, and the model without 2DSG$\to$3DSG in step 3, respectively.
As shown in Figure \ref{fig:sg_obj}, we can see both ST2I and SI2T performance decreases when removing any training steps.
Notably, the 2DSG$\to$2DSG step contributes the most to the final results. The essential 3D understanding function lies in the capability of the graph diffusion module, which should be largely enhanced via this step.

\vspace{-3mm}
\section{Conclusion}
\vspace{-2mm}

In this work, we study the ST2I and SI2T tasks under a dual learning framework.
We first propose the spatial-aware 3D scene graph to model the 3D feature which is essentially shared by ST2I and SI2T.
The 3DSG is constructed by being initialized with a 2D TSG/VSG and then evolved to the 3DSG by a graph diffusion model.
We then leverage the dual learning to enhance the 2DSG$\to$3DSG evolving by sharing clues of the dual 3DSG$\to$Image/Text process, through which the 3DSG diffusion model is adequately guided and then facilitates the whole ST2I and SI2T.
On the VSD dataset, our method shows great superiority in both SI2T and ST2I. 
Further analyses demonstrate how our dual learning method could capture 3D spatial structures and then help generate spatial-faithful images and texts.

Looking forward, there can be quite rich explorations in future research.
First, the graph diffusion model highly relies on the quality of 3DSG training data, which has significant impact to the final performance of our method.
Correspondingly, we in the future will explore the data augmentation methods to optimize the training.
Second, the evaluation for spatial understanding of ST2I and SI2T has not been fully explored in this work, i.e., with human evaluation instead.
Also, one promising direction for VSU is constructing multimodal large language models (MLLMs) \cite{fei2024multimodal,wu24next,fei2024vitron,wu2024towards,zhang2024omg,qian2024momentor,wu2024controlmllm}, especially for 3D spatial understanding \cite{chen2024ll3da}.

\vspace{-2mm}
\section*{Acknowledgements}

\vspace{-2mm}
This research is supported by CCF-Kuaishou Large Model Explorer Fund, 
Project of Future High-tech Video Intelligent Technology Innovation Center,
National Natural Science Foundation of China (NSFC) under Grant 62336008,
A*STAR AME Programmatic Funding A18A2b0046, RobotHTPO Seed Fund under Project C211518008,
and EDB Space Technology Development Grant under Project S22-19016-STDP.

{\small
\bibliographystyle{plainnat}
\bibliography{sample-base}
}


\newpage
\appendix



\section{Extension of Technical Details}

In this section, we introduce the specific details of our method which we cannot present in the main article due to the space limit.

\subsection{Discrete Diffusion}\label{app:ddiff}
We first introduce the discrete representation of continuous data.
Given data $\bm{x}$ in a continuous space, a vector quantized model (VQM) is employed to encode $\bm{x}$ to embedding vectors.
A VQM contains an encoder $E$, a decoder $G$ and a pre-trained codebook $\mathcal{Z}=\{\bm{c}_k\}_{k=1}^K \in \mathbb{R}^{K\times d}$, where $\mathcal{Z}$ has a finite number of embedding vectors, $K$ is the size of the codebook and $d$ is the code dimension.
Given input $\bm{x}$, we obtain a sequence of tokens $\bm{z}_q$ with the encoder $\bm{e}=E(\bm{x})$ and a quantizer that maps $\bm{e}$ to its closet codebook entry $\bm{c}_k$:
\begin{equation}\label{eq:1}
    \bm{z}_q=Q(\bm{e})=\mathop{argmin}_{\bm{}_k\in Z}\Vert \bm{e}-\bm{c}_k\Vert
\end{equation}
Correspondingly, given a quantified tokens $\bm{z}_q$, the decoder could faithfully reconstruct the data $\tilde{\bm{x}}=G(\bm{z}_q)$.

With this discrete representation, the discrete diffusion model can be applied to it.
Given the quantized data $\bm{z}_0$, the forward diffusion process gradually corrupts $\bm{z}_0$ through the Markov chain $q(\bm{z}_t|\bm{z}_{t-1})$, randomly replacing some tokens in $\bm{z}_{t-1}$.
After a fixed number of $T$ steps, the model outputs a sequence of increasingly noisy latent variables $\bm{z}_1,...,\bm{z}_T$, where $\bm{z}_T$ is the pure noise tokens.
In the reverse process, the model gradually denoise from $\bm{z}_T$ and reconstruct $\bm{z}_0$, by sampling from the distribution $q(\bm{z}_{t-1}|\bm{z}_t,\bm{z}_0)$ sequentially.
Specifically, for a token $z^i_0$ of $\bm{z}_0$, $z^i_0$ takes the index of one entry of codebook, i.e., $z^i_0\in \{1,...,K\}$.
The probabilities of the transition from $\bm{z}_{t-1}$ to $\bm{z}_t$ can be represented by a matrix $\bm{Q}_t(mn)=q(\bm{z}_t=m|\bm{z}_{t-1}=n)\in\mathcal{R}^{K\times K}$.
Then the forward diffusion process for $\bm{z}_t$ is:
\begin{equation}\label{eq:2}
    q(\bm{z}_t|\bm{z}_{t-1})=v^{\top}(\bm{z}_t)\bm{Q}_tv(\bm{z}_{t-1}),
\end{equation}
where $v(z)$ means the one-hot representation of $x$ in $K$ categories.
$\bm{Q}_tv(\bm{z}_{t-1})$ means the categorical distribution over $\bm{z}_t$.
Due to its Markov property, the $q(\bm{z}_t|\bm{z}_0)$ could be written as:
\begin{equation}\label{eq:3}
\begin{split}
    q(\bm{z}_t|\bm{z}_0)&=v^{\top}(\bm{z}_t)\left(\prod_1^t\bm{Q}_{t'}\right)v(\bm{z}_0)\\
    &=v^{\top}(\bm{z}_t)\overline{\bm{Q}}_tv(\bm{z}_0).
\end{split}
\end{equation}
By applying Bayes’ rule, we can compute the posterior $q(\bm{z}_{t-1}|\bm{z}_t,\bm{z}_0)$ as:
\begin{equation}
\begin{split}
    &q(\bm{z}_{t-1}|\bm{z}_t,\bm{z}_0)=\frac{q(\bm{z}_t|\bm{z}_{t-1},\bm{z}_0)q(\bm{z}_{t-1}|\bm{z}_0)}{q(\bm{z}_t|\bm{z}_0)}\\
    =&\frac{\left(v^{\top}(\bm{z}_t)\bm{Q}_{t-1}v(\bm{z}_{t-1})\right)\left(v^{\top}(\bm{z}_{t-1})\overline{\bm{Q}}_{t-1}v(\bm{z}_0)\right)}{v^{\top}(\bm{z}_t)\overline{\bm{Q}}_tv(\bm{z}_0)}.
\end{split}
\end{equation}
$\bm{Q}_t$ is usually defined as the a small amount of uniform noises and it can be formulated as:
\begin{equation}
    \bm{Q}_t=\left[\begin{array}{cccc}
       \alpha_t+\beta_t  & \beta_t & \cdots & \beta_t \\
       \beta_t & \alpha_t+\beta_t & \cdots & \beta_t \\
       \vdots & \cdots & \ddots & \vdots \\
       \beta_t & \beta_t & \cdots & \alpha_t+\beta_t \\
    \end{array}\right]
\end{equation}
where $\alpha_t\in[0,1]$ and $\beta_t=(1-\alpha_t)/K$, which means each token has a probability of $\alpha_t+\beta_t$ to remain the previous value at one step and has a probability of $K\beta_t$ to be sampled uniformly over all the $K$ categories.

Then a noise estimating network $p_{\theta}(\bm{z}_{t-1}|\bm{z}_t, \bm{c})$ is trained to approximate the conditional transit distribution $q(\bm{z}_{t-1}|\bm{z}_t,\bm{x}_0)$ with condition $\bm{c}$.
The network is trained to minimize the variational lower bound (VLB) \cite{DBLP:conf/icml/Sohl-DicksteinW15}.
\begin{equation}\label{eq:6}
    \mathcal{L}=D_{KL}(q(\bm{z}_{t-1}|\bm{z}_t,\bm{x}_0)\Vert p_{\theta}(\bm{z}_{t-1}|\bm{z}_t, \bm{c}))
\end{equation}

\subsection{Discrete Representation for 3DSG}
Following \cite{DBLP:journals/corr/abs-2403-16883}, given the 3DSG $\mathcal{G}$, we use a a GNN encoder to encode $G$ to a set of node embeddings $\mathcal{Z}={\bm{z}^G_i}$.
We subsequently apply a quantization operator, $\mathcal{F}$, on the continuous node embeddings and map them to fixed points within the same space.
The quantization operates independently on each dimension of the latent space, and the quantization of the $j^{th}$ dimension of $z_i$ embedding is given by:
\begin{equation}
    z^q_{ij} = \mathcal{F}(z^G_{ij},L_j)=\mathcal{R}(\frac{L_i}{2}\tanh z_{ij}), 1\leq j \leq d_Z
\end{equation}
where $\mathcal{R}$ is the rounding operator and $L_j$ is the number of quantization levels used for the $j^{th}$ dimension of the node embeddings. 
The quantisation operator maps any point in the original continuous latent space to a point from the set $\overline{Z} = {\overline{z}_i}$.
The discrete latent representation of graph $\mathcal{G}$ is the set $Z^G=\{z_i^q\}, z_i^q\in \overline{Z}$.

The quantization operator is permutation equivariant and so is the mapping from the initial graph G to its discrete representation in the latent space as long as the graph encoder $\mathcal{D}$ is permutation equivariant.
Thus for any $P$ permutation matrix the following equivariance relation holds:
\begin{equation}
    P^\top Z^G=\mathcal{F}(P^\top Z)=\mathcal{F}(\mathcal{D}(P^\top EP, P^\top X))
\end{equation}
Using an equivariant decoder $\mathcal{D}(Z^G)$, which will operate on the assumption of a fully-connected graph over the node embeddings $Z^G$, results in reconstructed graphs that are node permutation equivariant with the original input graphs.

\subsection{Aligned Image-Text-3DSG Data Construction.} 
We follow \cite{DBLP:journals/corr/abs-2312-11713} to take the 3D datasets Matterport3D (MP3D) \cite{DBLP:conf/3dim/ChangDFHNSSZZ17} 3DSSG \cite{DBLP:conf/cvpr/WaldDNT20} and CURB-SG \cite{DBLP:journals/corr/abs-2309-06635}, using the Hydra \cite{DBLP:journals/corr/abs-2305-07154} parser to generate the 3D scene graphs and then leverage large language models to refine it as the ground-truth.
Then we use ChatGPT to generate alignede spatial descriptions for the RGB images.

\subsection{Marginal Distribution Estimation for the Dual Learning Target}\label{app:dual}
Based on Eq. \ref{eq:dual}, the marginal distributions $p(x)$ and $p(y)$ can not be aqcired directly.
Thus we estimate these marginal distribution $p(x)$ and $p(y)$ with a surrogate distribution $\hat{p}(x)$ and $\hat{p}(y)$, by observing the target in the scope of the whole data.
For the text, we use a Transformer-based language model that is trained over the specific data to calculate the $\hat{p}(x)$ \cite{DBLP:conf/acl/SuHC19}.
For the image, we follow \cite{DBLP:conf/icml/0001WRZ22} to calculate $\hat{p}(y)=\prod_{t=1}^{m}px_i|x_{<i}$.
We serialize the image pixels as $x_i$ and use PixelCNN++ \cite{DBLP:conf/iclr/SalimansK0K17} to model this distribution.

\section{Detailed Experiment Settings}\label{app:ext_setting}
\subsection{Evaluation Metric Implication}
We employ Fréchet Inception Distance (FID) \cite{DBLP:conf/nips/SalimansGZCRCC16}, Inception score (IS) \cite{DBLP:conf/nips/HeuselRUNH17}, CLIP score \cite{DBLP:conf/emnlp/HesselHFBC21}, and GLIP \cite{DBLP:conf/cvpr/LiZZYLZWYZHCG22} used in \cite{DBLP:conf/aaai/FanC00YW23} to quantitatively evaluate the quality of the generated images.
Additionally, we introduce Triplet Recall (TriRec.) to measure the percentage of the correct relation triplet among all the relations. Technically, given a set of ground truth triplets (subject-predicate-object), denoted GT, and the TriRec. is computed as TriRec.=$|PT\cap GT |/|GT|$, where $PT$ are the relation triplets extracted from the generated images by a visual or textual SG parser.

\subsection{Human Evaluation Criterion}
We conduct a human evaluation to mainly focus on spatial understanding quality.

Then we design a 5-point Likert scale is designed as follows:
\begin{itemize}
    \item SI2T Spatial Accuracy: The sentences correctly describe the spatial relationship of the objects in the given image.
    \item SI2T Spatial Diversity: The generated sentences describe diversified spatial semantics.
    \item ST2I Spatial Accuracy; The image present the spatial description correctly. 
\end{itemize}
Each question will be answered by a number from 1 to 5, denoting “Strongly Disagree”, “Disagree”, “Neither disagree nor agree”, “Agree” and “Strongly agree”.
We generated 5 sentences for SI2T and 3 images for ST2I from 100 samples.

\section{Extended Experiment Results and Analyses}
\input{tables/app_res2}
\subsection{Comparison with More Methods}
We add more experiments of the latest T2I methods in table \ref{tab:app_res22}.
Comparing with the recent SD and CogView, our method exhibit leading performance with the similar scale.

\subsection{The Superiority of the Discrete Modeling}
To verify the superiority of the discrete modeling, we compare with the continuous diffusion backbone in table \ref{tab:app_diff}.
We find that the results of continuous backbone drops slightly on ST2I and SI2T while drops more on 3DSG generation, revealing that the discrete model show superiority for structure data modeling and further benefit the final performance.
\input{tables/app_diff}

\subsection{Can Non-diffusion SI2T models be used in our Dual Learning Framework?}
We use the diffusion based SI2T model to maintain the symmetry for diffusion based ST2I and 3DSG generation, so that their intermediate feature could be mutually enhanced.
Here we explore whether a vision-language model (VLM) based SI2T model could be integrated in our dual framwework.
We follow \cite{DBLP:conf/acl/Zhao00WZZC23} to use a OFA model as SI2T backbone.
We replace the $\bm{z}^Y_t$ in $p_{\psi}^{Graph}(\bm{z}^G_{t-1}|\bm{z}^G_{t},\bm{c}^Y\oplus \bm{z}^Y_t)$ to the OFA encoder output hidden states $\bm{z}^Y_{enc}$.
Then we train our model with the same training data and training strategy.
The results are shown in Table \ref{tab:app_si2t}, where we find the ST2I performance drops while the SI2T performance keep comparable.
This reveal the necessity of choosing diffusion based SI2T.

\input{tables/app_si2t}
\input{tables/app_efficiency}

\subsection{How the quality of 3DSG dataset influences the performance}
Instinctive, the quality of 3DSG training data has strong impact to the graph diffusion.
To verify this, we corrupt the gold 3DSG by random replace the node and edeges with a probability and compare the model in Figure \ref{fig:app_noise}.
We can see that with the noise increase, both the 3DSG and the final performance decrease sharply.
This means the graph diffusion model is sensitive to the 3DSG data quality.

\input{figures/app_sg_quality}

\subsection{Visualization of 3DSG Generation Process}
\begin{figure}
    \centering
    \includegraphics[width=\textwidth]{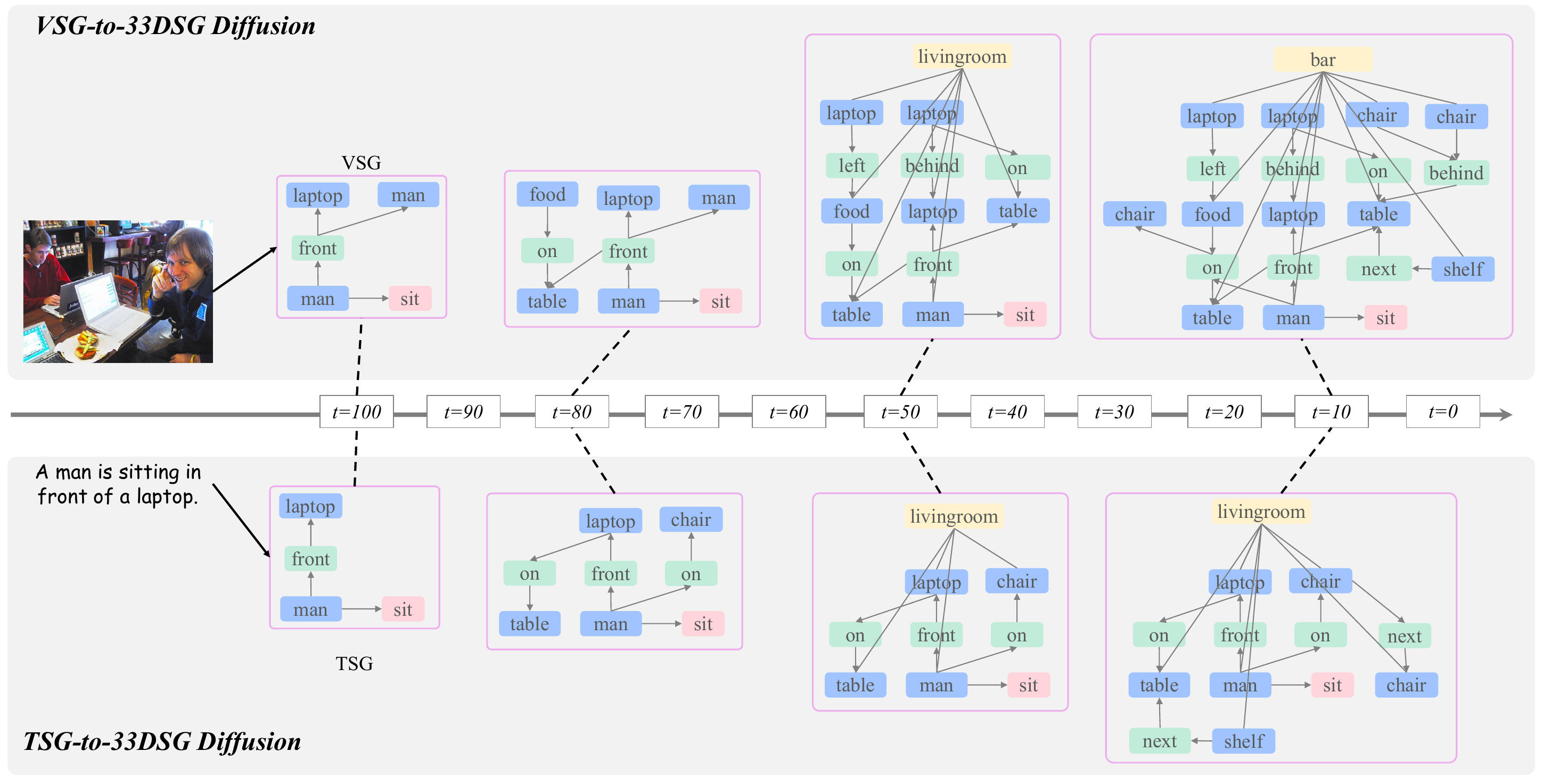}
    \caption{Visualization of 3DSG generation process from initial VSG and TSG.}
    \label{fig:app_graph_diff}
\end{figure}

In Figure \ref{fig:app_graph_diff}, we visualize the 3DSG generation process with two examples, where we sample several time steps and plot the generated 3DSG can be seen, compared with the initial VSG/TSG our graph diffusion is capable of enrich certain reasonable structures.

\subsection{Efficiency Analysis}
In table \ref{tab:app_time}, we give the computational complexity analysis as well as our training settings.

\subsection{More Cases}
\begin{figure}
    \centering
    \includegraphics[width=\textwidth]{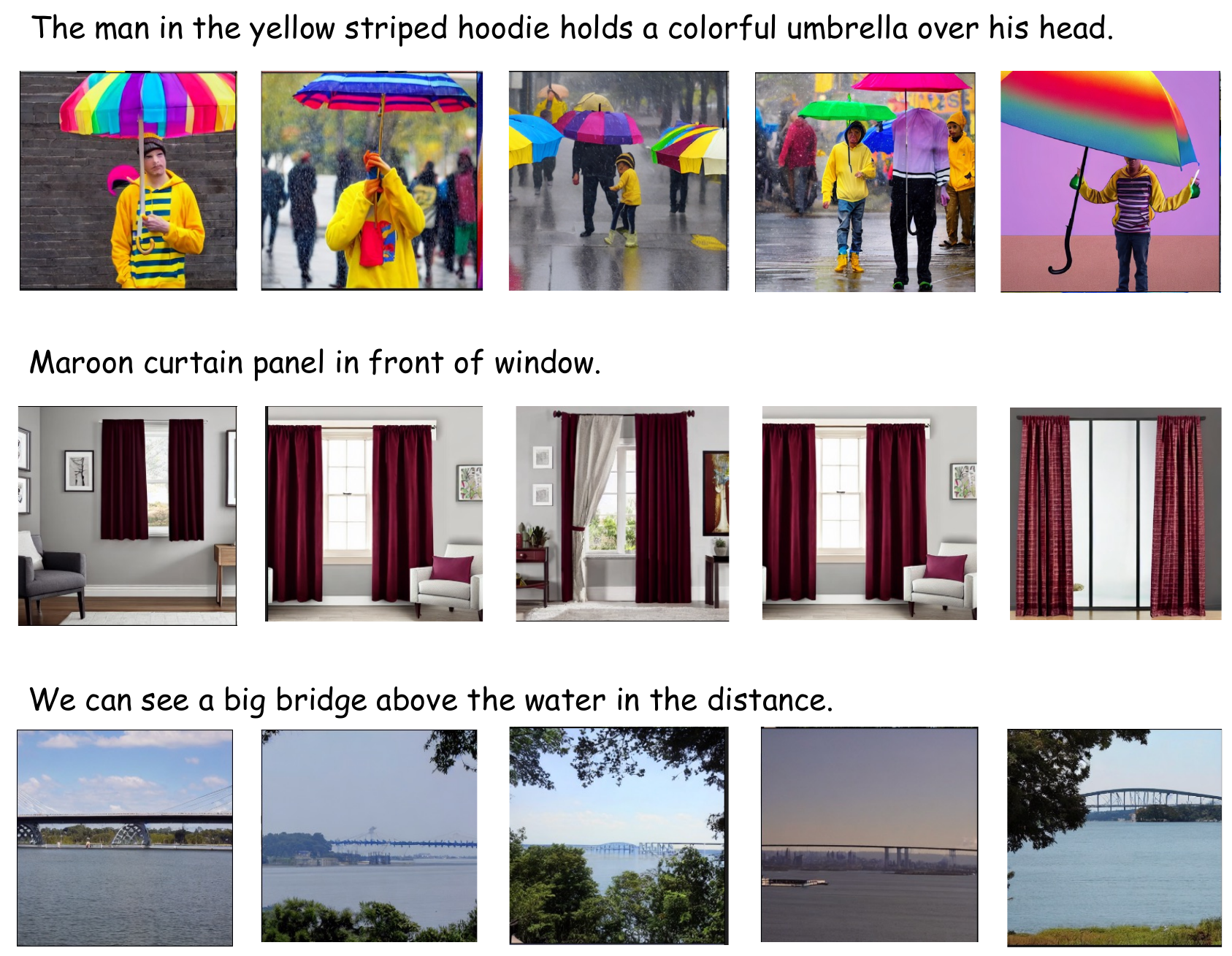}
    \caption{More cases of ST2I generated by SD$^3$.}
    \label{fig:app_case_t2i}
\end{figure}

\begin{figure}
    \centering
    \includegraphics[width=\textwidth]{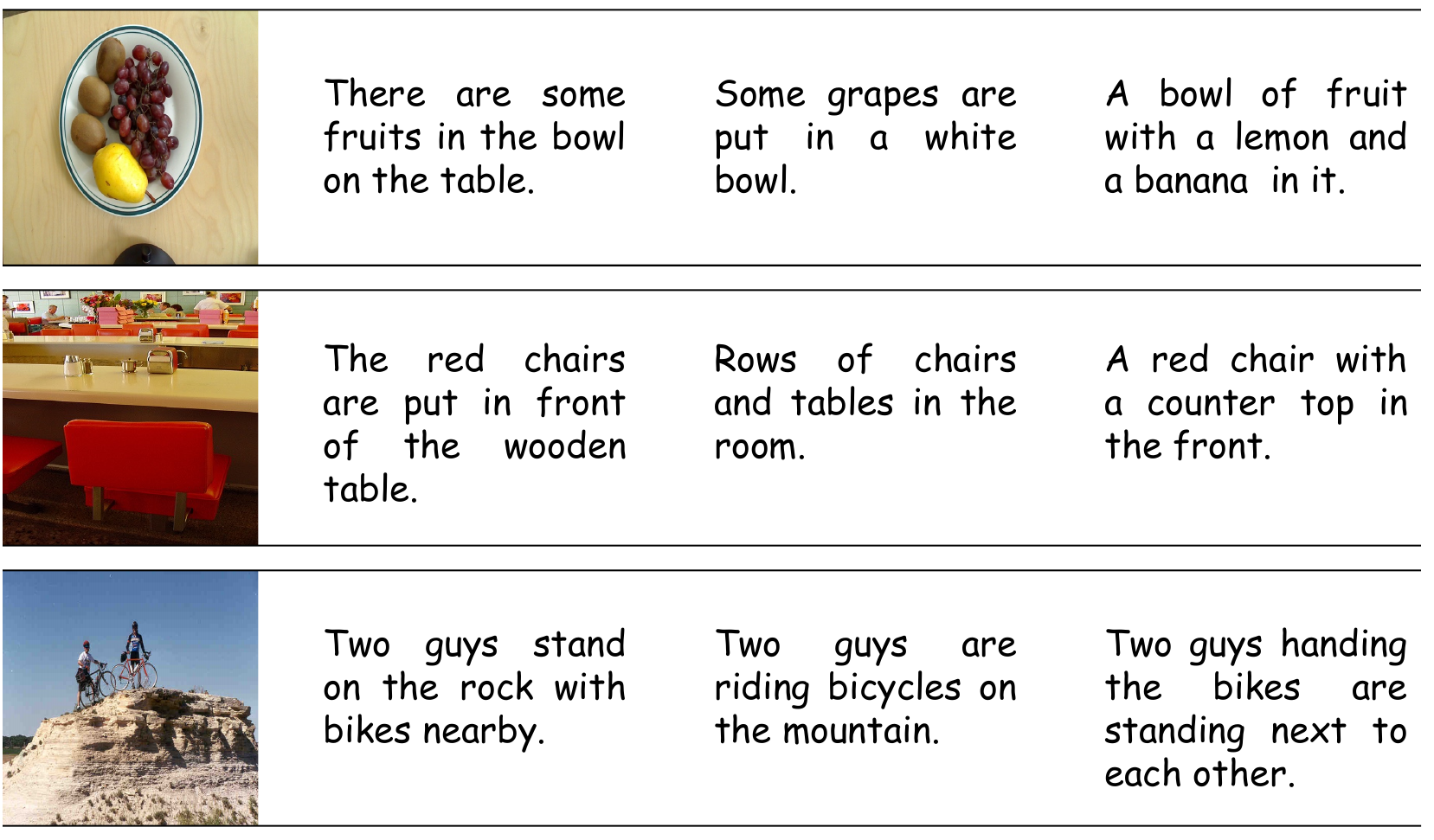}
    \caption{More cases of SI2T generated by SD$^3$.}
    \label{fig:app_case_i2t}
\end{figure}

Here we showcase more examples of the ST2I and SIT@ via SD$^3$ in Figure \ref{fig:app_case_t2i} and Figure \ref{fig:app_case_i2t}.

\clearpage


\end{document}

%% file: tables/3DSG.tex
\newenvironment{tabitemize}[1][]
 {\begin{itemize}[nosep,#1]}
 {\end{itemize}}

\begin{wraptable}[19]{r}{8cm}
\fontsize{9}{11}\selectfont
\linespread{0.7}\selectfont
\vspace{-6mm}
\caption{Comparison of 3DSG and 2DSG.}
\vspace{-3mm}
\label{tab:3DSG}
\begin{center}
\begin{tabular}{@{}>{\raggedright\arraybackslash}p{8cm-2\tabcolsep-0.5\arrayrulewidth}@{}}
\toprule
\multicolumn{1}{c}{\textbf{2D TSG}} \\
\midrule
\begin{tabitemize}[leftmargin=*]
\item \textbf{Objects:} Only key objects
\item \textbf{Attributes:} Any attributes
\item \textbf{Relationships:} Semantic relationships between objects
\end{tabitemize}
\\[-6pt]
\midrule[\heavyrulewidth]
\multicolumn{1}{c}{\textbf{2D VSG}} \\
\midrule
\begin{tabitemize}[leftmargin=*]
\item \textbf{Objects:} The front objects in one perspective
\item \textbf{Attributes:}
Visual related attributes
\item \textbf{Relationships:}
Simple 2D spatial relationships
\end{tabitemize}
\\[-6pt]
\midrule[\heavyrulewidth]
\multicolumn{1}{c}{\textbf{3DSG}} \\
\midrule
\begin{tabitemize}[leftmargin=*]
\item \textbf{Objects:} All the objects in the 3D scene; High-level spatial concepts.
\item \textbf{Attributes:}
Any attributes including 3D attributes such as pose and texture.
\item \textbf{Relationships:}
3D spatial relationships and subordinate relationships between spatial concepts.
\end{tabitemize}
\\[-6pt]
\bottomrule
\end{tabular}
\end{center}
\end{wraptable}

%% file: tables/main_res.tex
\begin{table*}[t]
\fontsize{8}{10}\selectfont
\setlength{\tabcolsep}{1.4mm}
\caption{
Main results on the 256 × 256-sized VSD dataset with 200 DDIM steps.
\textbf{Bold} numbers are the best and the \underline{Underline} numbers denote the best baselines.
Our results are averaged on five running with different seeds.
}
\vspace{-3mm}
\begin{center}
\begin{tabular}{clcccccccccc}

\toprule
\multicolumn{2}{c}{\multirow{2}{*}{}} & \multicolumn{5}{c}{\bf VSDv1} & \multicolumn{5}{c}{\bf VSDv2} \\
\cmidrule(r){3-7} \cmidrule(r){8-12}
& & \multicolumn{3}{c}{\bf \texttt{ST2I}} & \multicolumn{2}{c}{\bf \texttt{SI2T}} & \multicolumn{3}{c}{\bf \texttt{ST2I}} & \multicolumn{2}{c}{\bf \texttt{SI2T}}\\
\cmidrule(r){3-5} \cmidrule(r){6-7} \cmidrule(r){8-10} \cmidrule(r){11-12}
\multicolumn{2}{c}{} &\bf FID$\downarrow$ &\bf IS$\uparrow$ &\bf CLIP$\uparrow$ & \bf BLEU4$\uparrow$ &\bf SPICE$\uparrow$ & \bf FID$\downarrow$ &\bf IS$\uparrow$ &\bf CLIP$\uparrow$ & \bf BLEU4$\uparrow$ &\bf SPICE$\uparrow$ \\
\midrule

\multicolumn{6}{l}{$\bullet$ \textbf{T2I Baselines}} \\
& DALLE \cite{DBLP:conf/icml/RameshPGGVRCS21} & 32.55 & 17.01 & 62.16 & - & - & 28.52 & 21.18 & 64.58 & - & - \\
& Cogview \cite{DBLP:conf/nips/DingYHZZYLZSYT21} & 32.30 & 17.07 & 61.85 & - & - & 28.17 & 21.74 & 64.76 & - & - \\
& LAFITE \cite{DBLP:journals/corr/abs-2111-13792} & 30.73 & 24.39 & & - & - & 25.73 & 25.47 &  & - & - \\
& VQ-Diffusion \cite{DBLP:conf/cvpr/GuCBWZCYG22} &  18.34 & 20.58 & 63.42 &  - &  - & 15.66 & 24.75 &  66.30 & - & - \\
& Friodo \cite{DBLP:conf/aaai/FanC00YW23} & \underline{12.86} & \underline{25.92} & \underline{64.65} & - & - & \underline{11.41} & \underline{26.02} & \underline{67.01} & - & - \\
\midrule
\multicolumn{6}{l}{$\bullet$ \textbf{I2T Baselines}}\\
& 3DVSD \cite{DBLP:conf/acl/Zhao00WZZC23} & - & - & - & \underline{54.85} & \underline{68.76} & - & -& - & \underline{26.40} & \underline{46.97} \\
& MNIC \cite{DBLP:journals/corr/abs-1906-00717} & - & - & - & 34.21 & 66.87 & - & -& - & 20.01 & 43.88\\
& FNIC \cite{DBLP:journals/corr/abs-1912-06365} & - & - & - & 37.03 & 66.50 & - & -& - & 22.62 & 43.52\\
& DiffCap \cite{DBLP:journals/corr/abs-2305-12144} & - & - & - & 34.75 & 66.39 & - & -& - & 20.27 & 43.30\\
& DDCap \cite{DBLP:journals/corr/abs-2211-11694} & - & - & - & 37.93 & 67.10 & - & -& - & 23.14 & 44.07\\

\midrule
%
& Singleton  & 18.05 & 20.42 & 63.51 & 48.77 & 66.59 & 14.70 & 24.62 & 66.41 & 23.51 & 43.70 \\
& Singleton + 3D  & 12.56 & 26.92 & 65.62 & 50.05 & 67.20 & 10.43 & 25.62 & 67.29 & 25.37 & 45.13\\
& Vanilla Dual Learning  & 11.80 & 27.85 & 67.18 & 51.59 & 67.79 & 11.67 & 27.80 & 68.46 & 26.10 & 46.72\\
& \bf SD$^3$ (Ours) & \bf 11.04 & \bf 29.20 & \bf 68.31 & \bf 56.23 & \bf 68.02 & \bf 10.09 & \bf 29.76 & \bf 71.10 & \bf 27.63 & \bf 48.03 \\
\bottomrule
\end{tabular}
\label{tab:result}
\end{center}
\vspace{-3mm}
\end{table*}

%% file: tables/human.tex
\begin{wraptable}[8]{r}{5cm}
\fontsize{9}{10}\selectfont
\setlength{\tabcolsep}{3mm}
\vspace{-4mm}
\caption{
Human Evaluation for Spatial Accuracy.
}
\vspace{-2mm}
\begin{center}
\setlength{\tabcolsep}{3pt}
\begin{tabular}{clcccccccccc}

\hline
& \multicolumn{1}{c}{\bf \texttt{ST2I}} & \multicolumn{2}{c}{\bf \texttt{SI2T}} \\
\cmidrule(r){2-2} \cmidrule(r){3-4} \cmidrule(r){8-10} \cmidrule(r){11-12}
& SpaAcc. & SpaAcc. & Div. \\
\hline
3D$^3$ &  3.86 & 4.03 & 3.62 \\
Frido &  3.04 & - & - \\
3DVSD &  - & 3.03 & 2.76 \\

\hline
\end{tabular}
\label{tab:human}
\end{center}
\vspace{-3mm}
\end{wraptable}

%% file: tables/ana3.tex

\definecolor{c1}{RGB}{137,91,239}
\definecolor{c2}{RGB}{48,211,239}
\definecolor{c3}{RGB}{239,145,177}
\definecolor{c4}{RGB}{203,153,126}
\definecolor{c5}{RGB}{107,112,092}
\begin{minipage}{1\columnwidth}
\centering
\begin{minipage}{0.35\columnwidth}
\centering
\includegraphics[width=0.85\linewidth]{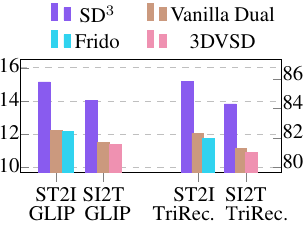}
\captionof{figure}{The evaluation of structure matching on VSDv2.}
\label{fig:q1}
\end{minipage}
\vspace{1mm}
\begin{minipage}{0.3\columnwidth}
\centering
\fontsize{8}{10}\selectfont
\setlength{\tabcolsep}{3.5mm}
\vspace{-3mm}
\captionof{table}{
Comparing the generated 3DSG with the gold 3GSD with TriRec.
}
\vspace{-2mm}
\begin{center}
\setlength{\tabcolsep}{4mm}

\begin{tabular}{lc}

\hline
Hydra (GT) & 100 \\
\hline
Initial TSG & 57.61 \\
Initial VSG & 24.20 \\
3D$^3$ w/o DiffPre. & 56.82 \\
3D$^3$ w/o Xfeat. & 65.71 \\
Full 3D$^3$ & 84.51 \\
\hline
\end{tabular}
\label{tab:q2}
\end{center}
\end{minipage}
\vspace{1mm}
\begin{minipage}{0.28\columnwidth}
    \centering
	

\includegraphics[width=0.9\linewidth]{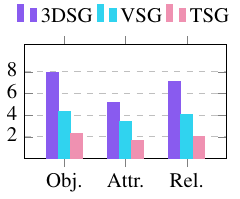}
\vspace{-2mm}
\captionof{figure}{The change of average number of SG nodes.}
\label{fig:q2_diff}
\end{minipage}
\end{minipage}

%% file: figures/obj_ablation.tex
\pgfplotsset{compat=1.7,every axis title/.append style={at={(0.5,-0.45)}, font=\fontsize{10}{1}\selectfont},every axis/.append style={xtick pos=left,ytick pos=left,tickwidth=1.5pt}}
\usetikzlibrary{matrix}
\usepgfplotslibrary{groupplots}
\usetikzlibrary{patterns,backgrounds}

\definecolor{c1}{RGB}{239,240,177}
\definecolor{c2}{RGB}{238,174,238}
\definecolor{c3}{RGB}{204,240,144}
\definecolor{c4}{RGB}{198,212,240}
\definecolor{c5}{RGB}{198,212,240}

\begin{wrapfigure}[11]{r}{8cm}
\centering
\includegraphics[width=1\linewidth]{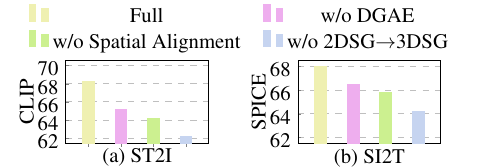}
\vspace{-3mm}
\caption{Comparison of different training strategy.}
\label{fig:sg_obj}
\end{wrapfigure}

%% file: tables/app_res2.tex
\begin{table}[htp]
\fontsize{8}{10}\selectfont
\setlength{\tabcolsep}{3mm}
\caption{
Comparison with latest T2I methods.
}
\begin{center}
\begin{tabular}{cccccccccccc}

\hline
& \multicolumn{3}{c}{\bf \texttt{ST2I}}\\
\cmidrule(r){2-4}
&\bf FID$\downarrow$ &\bf IS$\uparrow$ &\bf CLIP$\uparrow$ \\
\hline
SD-1.5	&20.12 &	26.35 &	64.79 \\
SDXL	&12.60 &	27.95 &	66.41 \\
CogView 3 &	13.28 &	28.63 &	67.29 \\
\bf Ours &\bf	11.04 &\bf	29.20 &\bf	68.31 \\

\hline
\end{tabular}
\label{tab:app_res22}
\end{center}
\vspace{-3mm}
\end{table}

%% file: tables/app_diff.tex
\begin{table}[htp]
\fontsize{8}{10}\selectfont
\setlength{\tabcolsep}{3mm}
\caption{
Comparison between discrete diffusion and continuous diffusion.
}
\begin{center}
\begin{tabular}{cccccccccccc}

\hline
& \bf \texttt{3DSG} & \multicolumn{3}{c}{\bf \texttt{ST2I}} & \multicolumn{2}{c}{\bf \texttt{SI2T}} \\
\cmidrule(r){2-2}\cmidrule(r){3-5} \cmidrule(r){6-7}
& \bf TriRec. &\bf FID$\downarrow$ &\bf IS$\uparrow$ &\bf CLIP$\uparrow$ & \bf BLEU4$\uparrow$ &\bf SPICE$\uparrow$ \\
\hline
w/ Descrete Diff (SD$^3$) & 84.51 & 10.09 & 29.76 & 71.10 & 27.63 & 48.03 \\
w/ Continuous Diff & 83.14 & 10.25 & 29.68 & 70.02 & 27.51 & 47.82 \\

\hline
\end{tabular}
\label{tab:app_diff}
\end{center}
\vspace{-3mm}
\end{table}

%% file: tables/app_si2t.tex
\begin{table}[htp]
\fontsize{8}{10}\selectfont
\setlength{\tabcolsep}{3mm}
\caption{
Comparison between diffusion and non-diffusion SI2T on VSDv2.
}
\begin{center}
\begin{tabular}{clcccccccccc}

\hline
& \multicolumn{3}{c}{\bf \texttt{ST2I}} & \multicolumn{2}{c}{\bf \texttt{SI2T}} \\
\cmidrule(r){2-4} \cmidrule(r){5-6}
&\bf FID$\downarrow$ &\bf IS$\uparrow$ &\bf CLIP$\uparrow$ & \bf BLEU4$\uparrow$ &\bf SPICE$\uparrow$ \\
\hline
w/ Diff-SI2T (SD$^3$)& 10.09 & 29.76 & 71.10 & 27.63 & 48.03 \\
w/ OFA-SI2T & 11.20 & 28.56 & 69.62 & 27.40 & 47.91 \\

\hline
\end{tabular}
\label{tab:app_si2t}
\end{center}
\vspace{-3mm}
\end{table}

%% file: tables/app_efficiency.tex
\begin{table}[ht]
\fontsize{8}{10}\selectfont
\setlength{\tabcolsep}{3mm}
\caption{
Efficiency analysis for each step.
}
\begin{center}
\begin{tabular}{lcccccccccc}

\hline
&\bf Training Parameters &\bf Training Time \\
\hline
Step-1 DGAE & 110M & 1 hours\\
Step-2 Spatial Alignment &	127M & 1.5 hours \\
Step-3 2DSG$\rightarrow$3DSG Diffusion Training &	350M &  12 hours \\
Step-4 Overall Training & 1.1B & 20 hours\\
\hline
\end{tabular}
\label{tab:app_time}
\end{center}
\vspace{-3mm}
\end{table}

%% file: figures/app_sg_quality.tex
\pgfplotsset{compat=1.7,every axis title/.append style={at={(0.5,-0.45)}, font=\fontsize{10}{1}\selectfont},every axis/.append style={xtick pos=left,ytick pos=left,tickwidth=1.5pt}}
\usetikzlibrary{matrix}
\usepgfplotslibrary{groupplots}
\usetikzlibrary{patterns,backgrounds}

\definecolor{c1}{RGB}{241,169,64}
\definecolor{c2}{RGB}{238,174,238}
\definecolor{c3}{RGB}{53,218,247}
\definecolor{c4}{RGB}{193,236,217}
\definecolor{c5}{RGB}{107,112,092}

\begin{figure}[htp]
\vspace{-5mm}
\centering
\begin{tikzpicture}
\begin{axis}[
	smooth,
	ytick = {0,20, 40, 60, 80, 100},
	yticklabels={0,20, 40, 60, 80, 100},
	ymin=0,ymax=105,
	ylabel = $\frac{\text{Noisy}}{\text{No Noise}}$ \scalebox{0.7}{(\%)},
	y tick label style = {yshift=-0.25em, text height=0ex,font=\small},
    x label style = {font=\small},
    xlabel = {Noise Rate (\%)},
    y label style = {yshift=-0.5em, font=\large,align=center},
    axis x line*=bottom,
	axis line style={-},
    axis y line*=left,
	axis line style={-},
	enlargelimits=0.05,
    every node near coord/.append style={black, font=\small, opacity=0.7, yshift=-0.0em,xshift=0.0em },
	legend style={at={(-0.11,1.01)},anchor=south west, draw=none, legend columns=-1, font=\small},
	xticklabels={0,20,40,60,80},
	xtick={1,2,3,4,5},
	ymajorgrids,
	xmax=5.5, xmin=0.5,
	x tick label style = {yshift=0.05em, align=center,font=\small},
	title style={yshift=-0.9em,font=\small},
	width = 8cm,
	height = 4cm,
	]
	\addplot[c1, thick, mark=*] coordinates
	{
		(1,  100)
            (2, 65)
            (3, 43)
            (4, 21)
            (5, 5)
	};\addlegendentry{SI2T}

        \addplot[c2, thick, mark=square] coordinates
	{
		(1, 100)
            (2, 75)
            (3, 63)
            (4, 41)
            (5, 23)
	};\addlegendentry{ST2I}
	
        \addplot[c3, thick, mark=star] coordinates
	{
		(1, 100)
            (2, 41)
            (3, 5)
            (4, 0)
            (5, 0)
	};\addlegendentry{3DSG}

\end{axis}

	
\end{tikzpicture}
\caption{The evaluation of structure matching on the gold 3DSSG dataset with noise.}
\label{fig:app_noise}
\vspace{-3mm}
\end{figure}